\documentclass{article}

\usepackage{microtype}
\usepackage{graphicx}
\usepackage{subfigure}
\usepackage{booktabs} % for professional tables

\usepackage{amsmath}
\usepackage{amssymb}
\usepackage{multirow}
\usepackage{caption}
\usepackage{tikz}
\usepackage{hyperref}

\usepackage[accepted]{icml2018}

\icmltitlerunning{RadialGAN: Leveraging multiple datasets to improve target-specific predictive models using GANs}

\begin{document}

\twocolumn[
\icmltitle{RadialGAN: Leveraging multiple datasets to improve target-specific predictive models using Generative Adversarial Networks}



\icmlsetsymbol{equal}{*}

\begin{icmlauthorlist}
\icmlauthor{Jinsung Yoon}{to}
\icmlauthor{James Jordon}{goo}
\icmlauthor{Mihaela van der Schaar}{to,goo,ati}
\end{icmlauthorlist}

\icmlaffiliation{to}{University of California, Los Angeles, CA, USA}
\icmlaffiliation{goo}{University of Oxford, UK}
\icmlaffiliation{ati}{Alan Turing Institute, UK}

\icmlcorrespondingauthor{Jinsung Yoon}{jsyoon0823@gmail.com}
\icmlkeywords{Machine Learning, ICML}

\vskip 0.3in
]



\printAffiliationsAndNotice{}  % leave blank if no need to mention equal contribution

\begin{abstract}
Training complex machine learning models for prediction often requires a large amount of data that is not always readily available. Leveraging these external datasets from related but different sources is therefore an important task if good predictive models are to be built for deployment in settings where data can be rare. In this paper we propose a novel approach to the problem in which we use multiple GAN architectures to learn to translate from one dataset to another, thereby allowing us to effectively enlarge the target dataset, and therefore learn better predictive models than if we simply used the target dataset. We show the utility of such an approach, demonstrating that our method improves the prediction performance on the target domain over using just the target dataset and also show that our framework outperforms several other benchmarks on a collection of real-world medical datasets.
\end{abstract}

\section{Introduction}\label{sect:intro}
Modern machine learning methods often require large amounts of data in order to learn the large number of parameters that define them efficiently, this may be because the model itself is complex (such as a multi-layer perceptron) and/or because the dimensions of the input data are large (as is often the case in the medical setting \cite{ahmed_tbme,yoon_plosone}). On the other hand, large datasets for a specific task such as counterfactual estimation and survival analysis may not be readily available \cite{yoon_ganite,lee_aaai}. Equally, it might be the case that it is desirable to learn a model that performs well on a specific, potentially small, sub-population. For example, in the medical setting it is important that models to be deployed for use in hospitals perform well on each hospital's patient population (rather than simply performing well across the entire population) \cite{sameproblem,targetonly,yoon_jbhi}. Learning an accurate model using the limited data from a single hospital can be difficult due to the lack of data and it is therefore important to leverage data from other hospitals, while avoiding {\em biasing} the learned model.

There are two main challenges presented when attempting to utilize data from multiple sources: feature mismatch and distribution mismatch. Feature mismatch refers to the fact that even among datasets drawn from the same field (such as medicine), the features that are actually recorded for each dataset may vary. It can often be the case, for example, that the practices in different hospitals lead to different features being measured \cite{yoon_jbhi}. The challenge this poses is two-fold - we need to deal with the fact that the "auxiliary" hospitals' datasets do not contain all the features that have been measured by the target hospital and moreover we leverage the information contained in the features that are measured by the auxiliary hospitals but not the target hospital.

Distribution mismatch refers to the fact that the patient population across two hospitals may vary. For instance, it may be the case that hospitals located in wealthier areas serve wealthier patients and that because of a patient's wealth, they are likely to have received a different standard of care in the past, therefore one might expect that patients in this hospital are more likely to exhibit ``healthier" characteristics, whereas a hospital in a poorer area is more likely to have an average patient exhibit ``sicker" characteristics. In order to utilize the datasets from several hospitals to construct a hospital-specific predictive model, we need to deal with the distribution mismatch between the source and target datasets. Otherwise, the learned predictive model will be {\em biased} and could perform poorly in the targeted setting. There is a wealth of literature that focuses solely on this problem, working under the assumption that all domains contain the same features. This paper provides a natural solution to both and is therefore solving a more general problem than frameworks tackling only distribution mismatch.

In this paper, we propose a novel approach for utilizing datasets from multiple sources to effectively enlarge a target dataset. This has strong implications for learning target-specific predictive models, since an enlarged dataset allows us to train more complex models, thus improving the prediction performance of such a model. We generalize a variant of the well known generative adversarial networks (GAN) \cite{standardgan}, namely CycleGAN \cite{cyclegan}. The proposed model, which we call RadialGAN, provides a natural solution to the two challenges outlined above and moreover is able to {\em jointly} perform the task for each dataset (i.e. it simultaneously solves the problem for each dataset as if it were the target). We use multiple GAN architectures to translate the patient information from one hospital to another, leveraging the adversarial framework to ensure that the learned translation respects the distribution of the target hospital. To learn multiple translations efficiently and simultaneously, we introduce a latent space through which each translation occurs. This has the added benefit of naturally addressing the problem of feature mismatch - all samples are mapped into the same latent space.

We evaluate the proposed model against various state-of-the-art benchmarks including domain translation frameworks such as CycleGAN \cite{cyclegan} and StarGAN \cite{stargan} using a set of real-world datasets. We use the prediction performance of two different predictive models (logistic regression and multi-layer perceptrons) to measure the performance of the various translation frameworks.

\subsection{Related Works}\label{sect:related_work}
\textbf{Utilizing datasets from multiple sources: }Several previous studies have addressed the problem of utilizing multiple datasets to aid model-building on a specific target dataset. \cite{targetonly,modeltransfer1,modeltransfer2,modeltransfer3} address this problem in the setting where all datasets have identical features and provide no solution to the feature mismatch problem described in the introduction. On the other hand, \cite{sameproblem} addresses only the problem of feature mismatch (specifically in the medical setting) and does not address the distributional differences that exist across the datasets. In spirit, our paper is most similar to \cite{sameproblem} as both are attempting to provide methods for utilizing multiple medical datasets that each contain a different set of features.

\textbf{Conditional GAN: }The conditional GAN framework presented in \cite{conditionalgan} provides an algorithm for learning to generate samples conditional on a discrete label. We draw motivation from this approach when we define our solution to the problem when the label of interest lies in a discrete space. However, we note (and demonstrate in Fig. \ref{fig:target_only_gan}) that a conditional GAN has limited applications in improving prediction performance - we cannot hope to generate samples that contain further information about the data than already contained in the dataset used to train the data generating process if the input is simply random noise.

\textbf{Pairwise dataset utilization: } Adversarially Learned Inference (ALI) \cite{ALI1,ALI2} proposes a framework for learning mappings between two distributions. It matches the joint distribution of sample and mapped sample of one distribution to sample and mapped sample of the other. However, due to the lack of restrictions on the conditional distributions, the learned functions do not satisfy cycle-consistency, that is, when mapping from one distribution to the other and then back again, the output is not the same as the original input. In other words, the characteristics of the original sample are ``lost in translation".

Adversarially Learned Inference with Conditional Entropy (ALICE) \cite{ALICE} is an extension of ALI that learns mappings which satisfy both reversibility (i.e. that the distribution of mapped samples match those of the other distribution in both directions) and cycle-consistency. More specifically, ALICE introduces the conditional entropy loss in addition to the adversarial loss to restrict the conditionals. This conditional entropy loss forces cycle-consistency.

CycleGAN \cite{cyclegan} and DiscoGAN \cite{discogan} propose further frameworks for estimating cycle-consistent and reversible mappings between two domains. Using explicit reconstruction error in place of the conditional entropy, CycleGAN and DiscoGAN ensure cycle-consistency. However, ALICE, CycleGAN and DiscoGAN are not scalable to multiple domains because the number of mappings to be learned is $M(M-1)$ when we have $M$ datasets. Furthermore, each pair of mappings only utilizes the two corresponding datasets when learning their parameters, on the other hand our framework is able to leverage all datasets when learning to map between a pair of datasets.

\textbf{Multi-domain translation: }StarGAN \cite{stargan} proposes a framework for multi-domain translation that is scalable to multiple domains by using a single generator (mapping) that takes as an additional input the target domain that the sample is to be mapped to. This framework utilizes all datasets to optimize the single generator. However, it only applies when there is no feature mismatch between domains. Moreover, the restrictive nature of modeling all mapping functions as a single network may create problems when the mapping functions between different pairs of domains are significantly different.

\section{Motivation} \label{sec:motiv}

\begin{figure*}[!htb]
	\centering
	\begin{minipage}{0.19\textwidth}
		\centering
		\includegraphics[width=.9\textwidth]{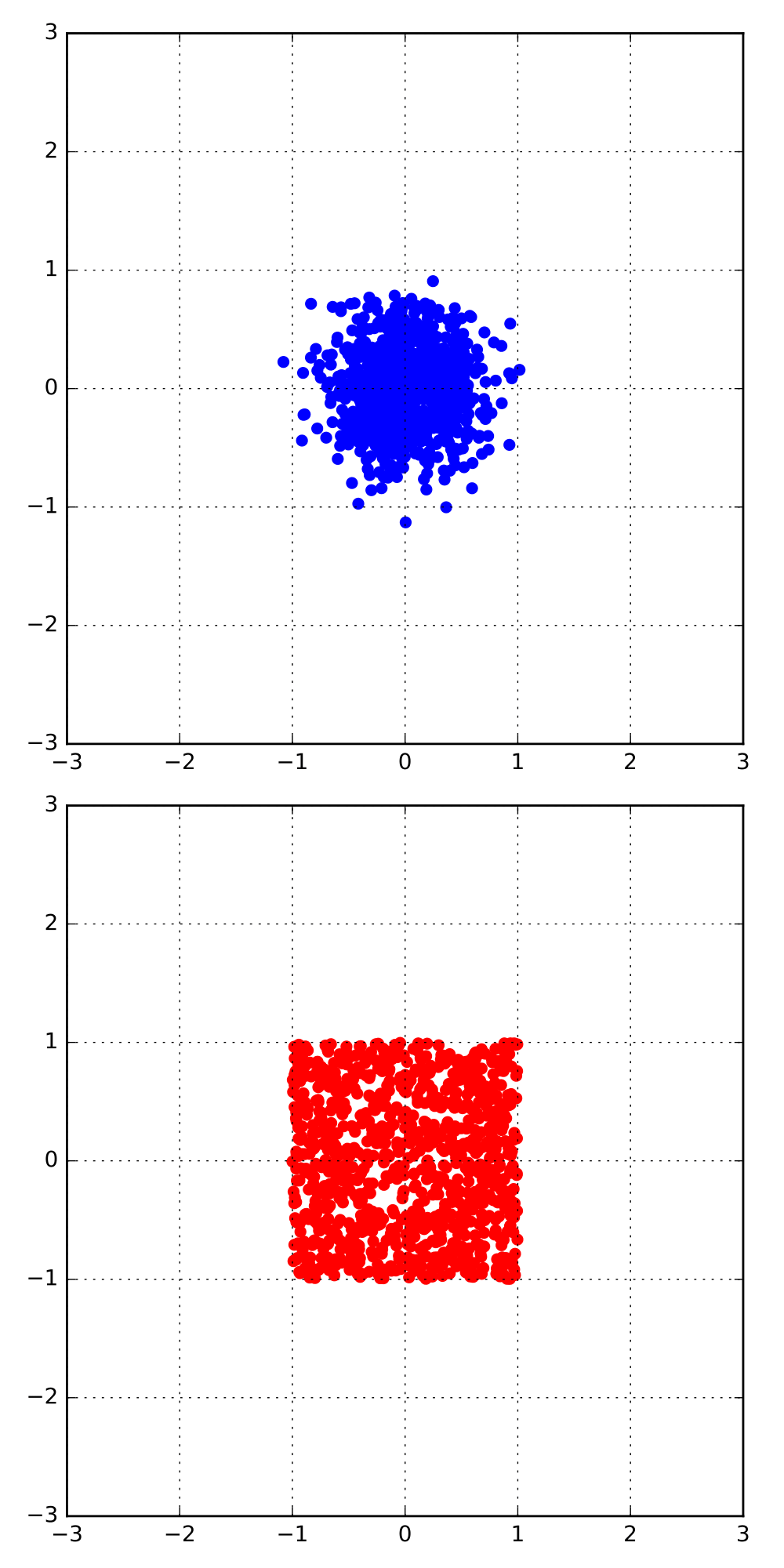}
		\captionsetup{labelformat=empty}
		\caption*{(a) Initial distribution}
	\end{minipage}
	\textcolor{black}{\vrule width 2pt}
	\begin{minipage}{0.19\textwidth}
		\centering
		\includegraphics[width=.9\textwidth]{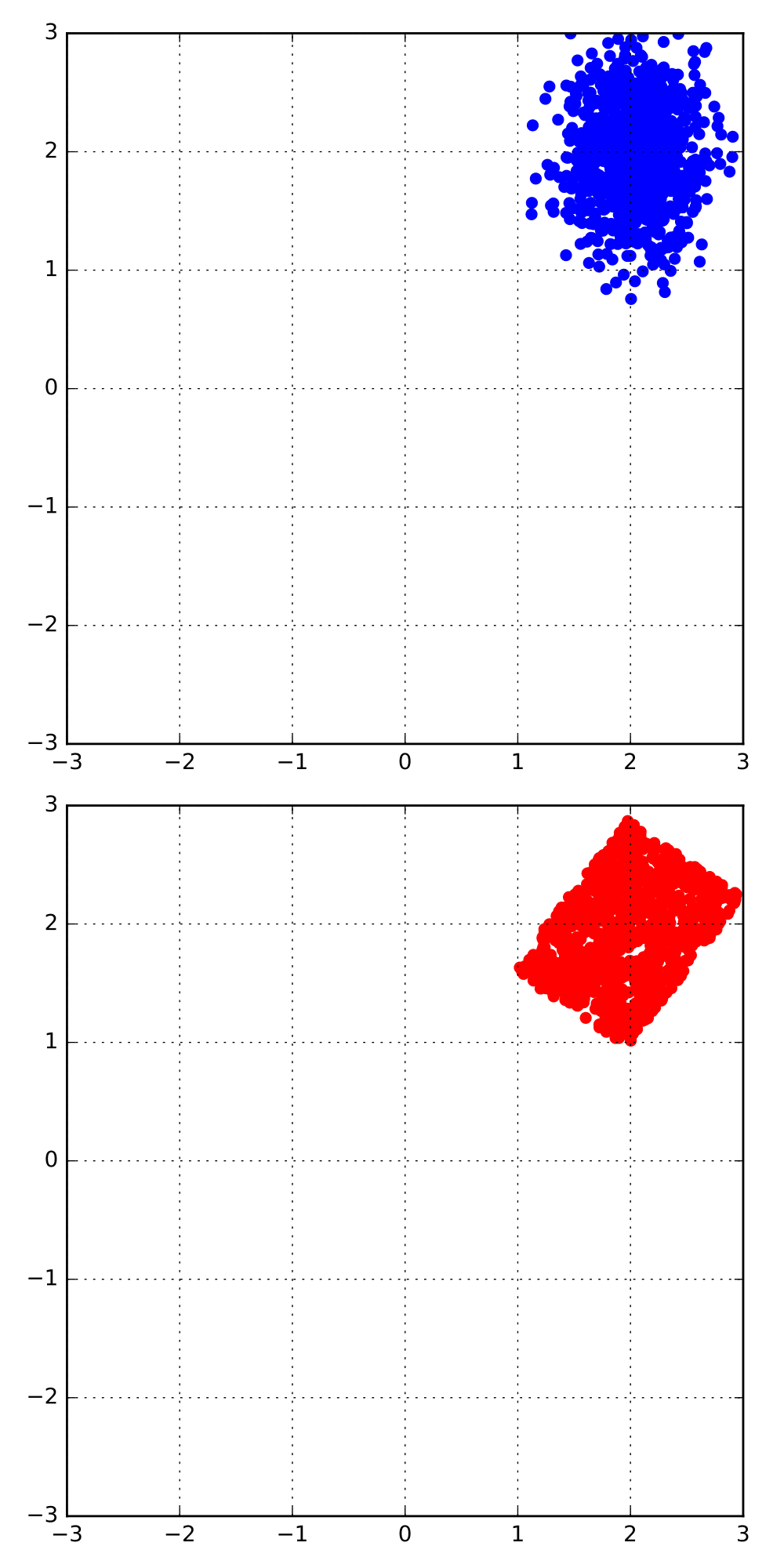}
		\captionsetup{labelformat=empty}
		\caption*{(b) linear model}
	\end{minipage}
	\begin{minipage}{0.19\textwidth}
		\centering
		\includegraphics[width=.9\textwidth]{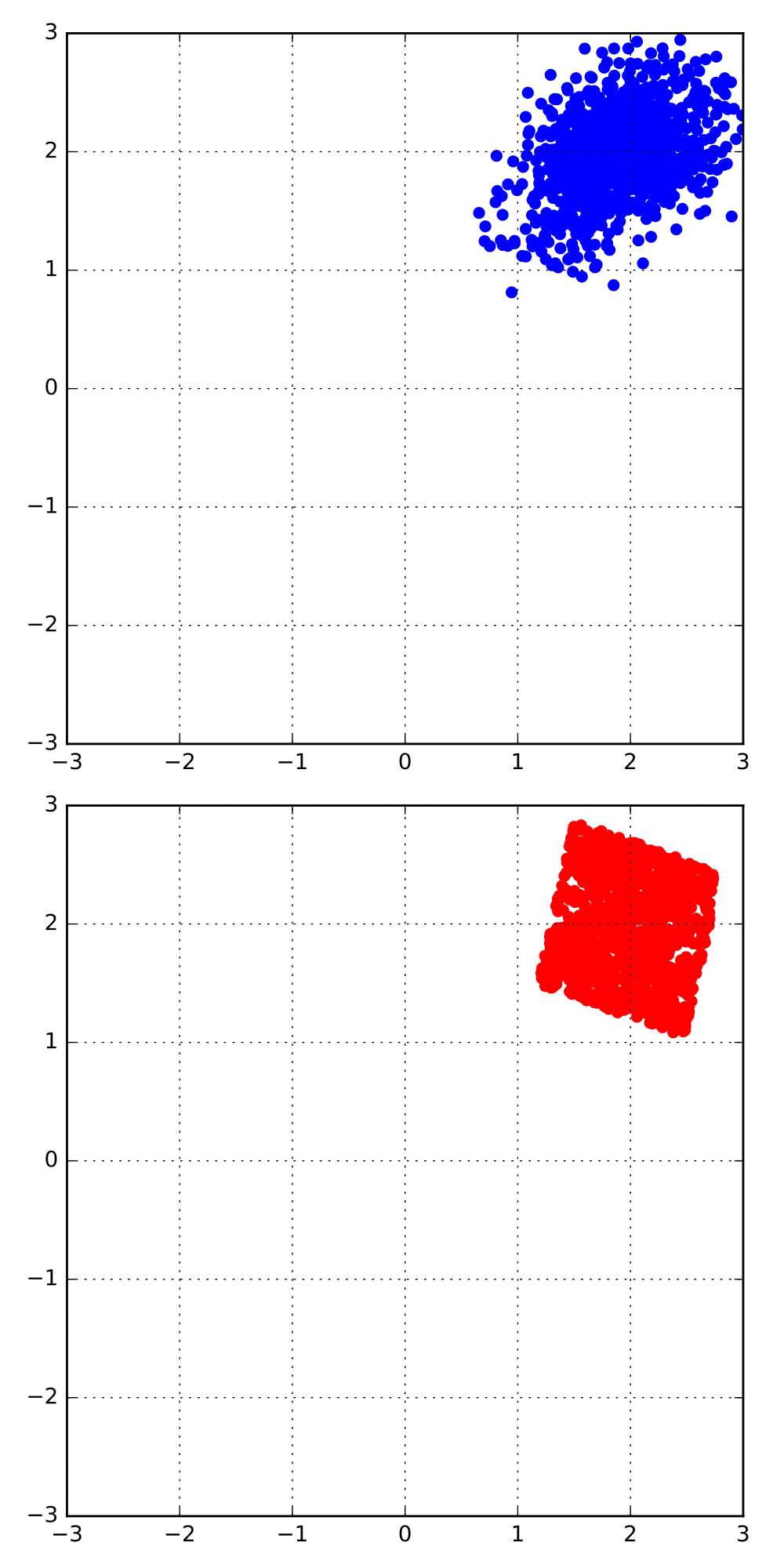}
		\captionsetup{labelformat=empty}
		\caption*{(c) one-layer perceptron}
	\end{minipage}
	\begin{minipage}{0.19\textwidth}
		\centering
		\includegraphics[width=.9\textwidth]{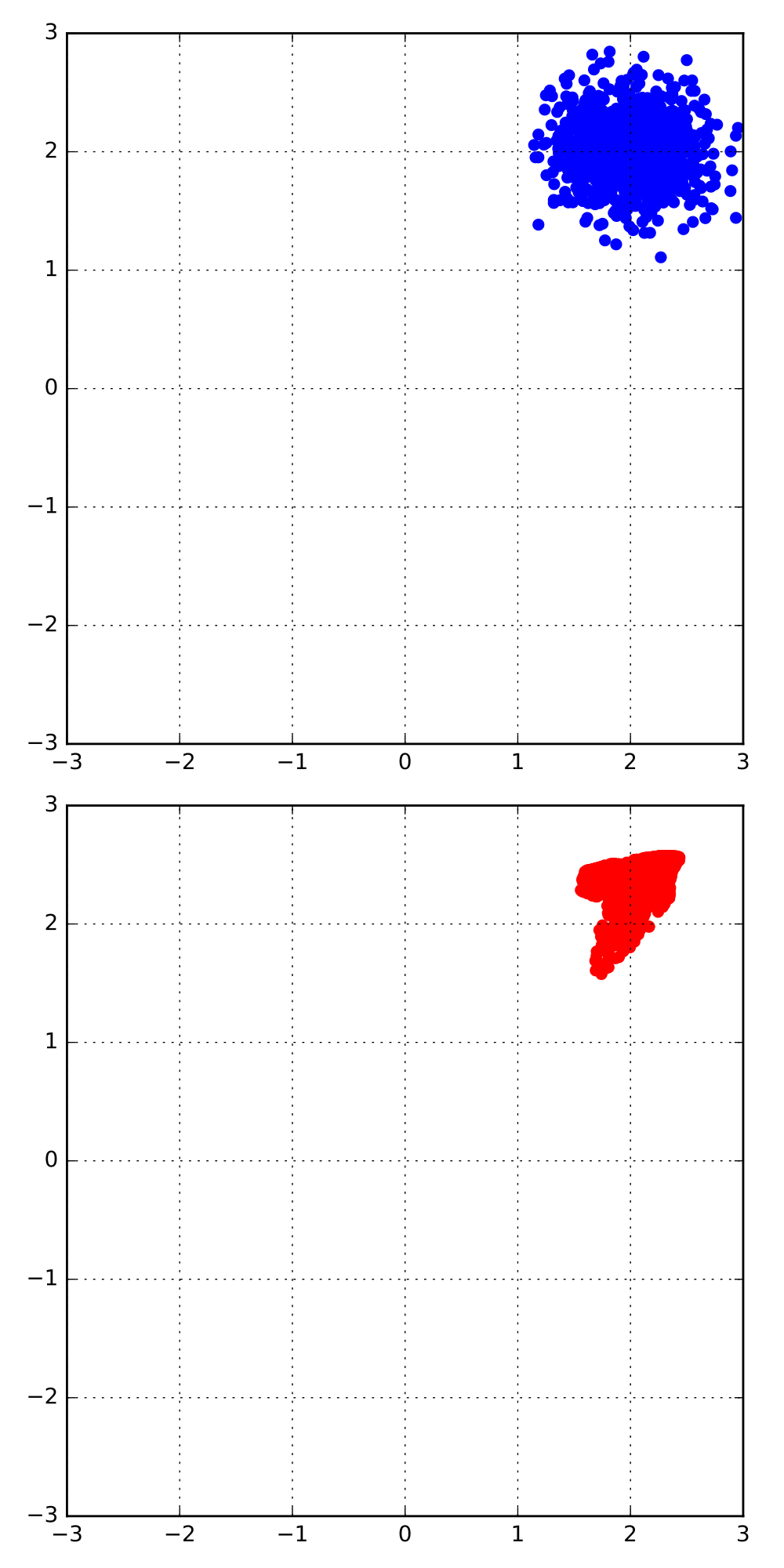}
		\captionsetup{labelformat=empty}
		\caption*{(d) two-layer perceptron}
	\end{minipage}
	\textcolor{black}{\vrule width 2pt}
	\begin{minipage}{0.19\textwidth}
		\centering
		\includegraphics[width=.9\textwidth]{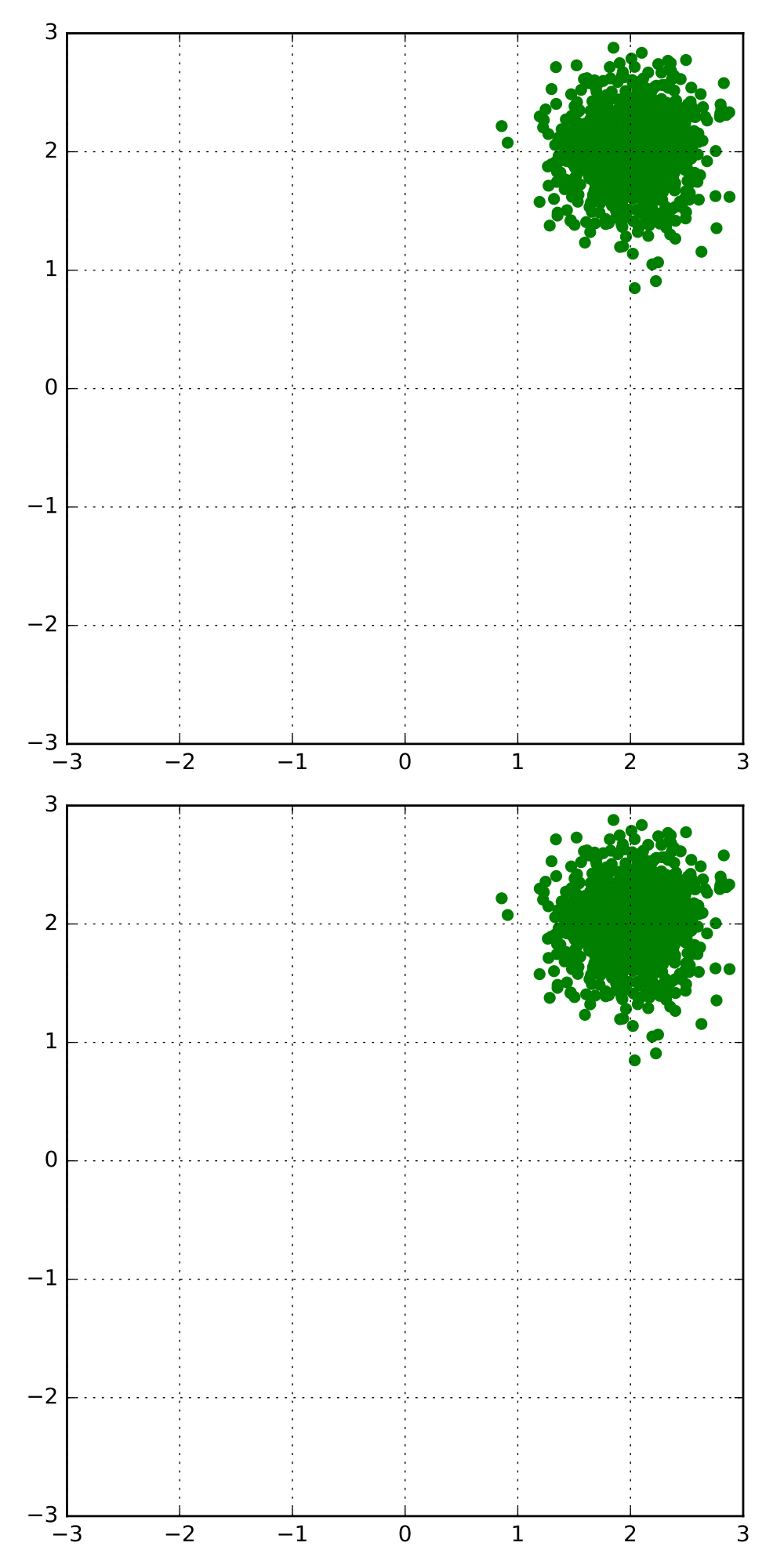}
		\captionsetup{labelformat=empty}
		\caption*{(e) target distribution}
	\end{minipage}
	\caption{Learning to map to the same target distribution for two different initial distributions (upper: Gaussian, lower: uniform)}
	\label{fig:motivation1}
\end{figure*}

To understand our solution to this problem we consider a simple toy example. Suppose that $X_1 \sim \mathcal{N}(\mathbf{0}, I_2)$, $X_2 \sim \mathcal{N}(\mu, I_2)$ and that $Z \sim \mathcal{U}([-1, 1]^2)$, so that $X_1, X_2, Z \in \mathbb{R}^2$. Then there is a ``simple" function $f_1$ such that $f_1(X_1) \overset{d.}{=} X_2$, namely, $f_1(\mathbf{x}) = \mathbf{x} + \mu$. On the other hand, an $f_2$ such that $f_2(Z) \overset{d.}{=} X_2$ is not as ``simple". If for example we attempt to learn $f_i$ using linear regression, then our approximation of $f_1$ will be far better than our approximation of $f_2$, primarily because the function $f_1$ lies in the space of linear models, and $f_2$ does not. Similarly, if we model $f$ as a neural net, then a larger capacity of neural net will be required in order to learn a suitable $f_2$ than to learn a suitable $f_1$. In particular, in order to learn more complex functions, we require a network with a larger capacity, however, in order to train such a network, we require more training samples. Note that learning $f_i$ is precisely what the GAN framework attempts to do, with $f_1$ being learned in the case that the GAN is given Gaussian noise, and $f_2$ being learned in the case that it is given uniform noise. In particular, the GAN framework can be very sensitive to the type of input noise it is given.

Fig. \ref{fig:motivation1} demonstrates this toy example for $\mu = (2, 2)$. As can be seen in the top row of the figure, when the initial and target distributions have similar shapes, the GAN framework is capable of learning the function even when the models are restricted to being low-capacity. On the other hand, the bottom row demonstrates that, even given the moderate capacity of a two-layer perceptron, the shape of the initial distribution can make learning to map to the target distribution difficult.

Note that the preceding discussion implies that (at least intuitively) if two random variables have similarly ``shaped" distributions, then a mapping between them will be ``simpler". In particular, if we have access to a random variable we believe to be very similar to our target, we can expect that generating samples of our target using this auxiliary variable will be easier than generating samples of our target using random noise. This motivates our approach to the transfer learning problem. We use the auxiliary datasets as the noise for a GAN framework, relying on the fact that the complex shape of a hospital's patient distribution will be better matched by {\em another} hospital's patient distribution, than by random (Gaussian) noise.

\section{RadialGAN}
Suppose that we have $M$ spaces $\mathcal{X}^{(1)}, ..., \mathcal{X}^{(M)}$, and that for each $i$, $X^{(i)}$ is a random variable taking values in $\mathcal{X}^{(i)}$. Suppose further that $Y$ is a random variable taking values in some label space $\mathcal{Y}$. Suppose that we have $M$ datasets $\mathcal{D}_1, ..., \mathcal{D}_M$ with $\mathcal{D}_i = \{(x^{(i)}_j, y^i_j)\}_{j = 1}^{n_i}$ where $(x^{(i)}_j, y^i_j)$ are i.i.d. realizations of the pair $(X^{(i)}, Y)$ and $n_i$ is the total number of realizations (observations) in dataset $i$.

For example, each $\mathcal{X}^{(i)}$ may correspond to the space of features (such as age, weight, respiratory rate etc.) that the $i$th hospital (of $M$) records, $X^{(i)}$ to a patient from the $i$th hospital and $Y$ might be 1-year mortality of the patient. The dataset $\mathcal{D}_i$ therefore contains observations of several patients from hospital $i$.

Our goal is to learn $M$ predictors, $f_1, ..., f_M$ (with $f_i : \mathcal{X}^{(i)} \to \mathcal{Y}$) such that $f_i$ estimates $Y$ for a given realization of $X^{(i)}$ and we want to utilize the datasets $\{\mathcal{D}_j : j \neq i\}$ (as well as $\mathcal{D}_i$) in learning $f_i$.

\begin{figure}[t!]
	\centering
	\includegraphics[width=0.4\textwidth]{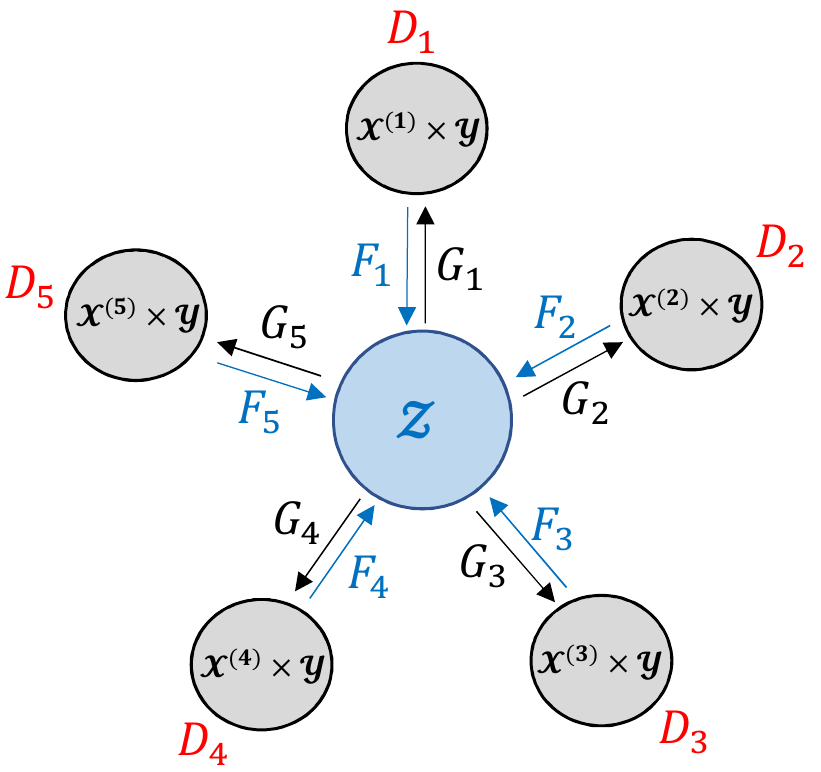}
	\caption{RadialGAN Structure - $\mathcal{Z}$: latent space, $\mathcal{X}^{(i)} \times \mathcal{Y}$: $i$th domain, $G_i,F_i,D_i$: Decoder, Encoder, and Discriminator of $i$th domain. The $i$th domain is translated to the $j$th domain via $\mathcal{Z}$ using $F_i$ and $G_j$.}
	\label{fig:structure}
\end{figure}

Motivated by the preceding discussion, we do this by using the datasets $\{\mathcal{D}_j : j \neq i\}$ as input to a GAN generator, noting that the datasets collected within the same field (e.g. medical datasets) are likely to be observing random variables with similar shapes. We provide two slightly different instances of our framework depending on whether the label, $Y$, is discrete or continuous. An overall picture of both architectures is captured in Fig. \ref{fig:structure}.

\vspace{-2mm}

For what follows, we introduce a space $\mathcal{Z}$, that we refer to as the {\em latent} space, and let $\alpha_{ij} = \frac{n_j}{\sum_{k \neq i} n_k}$. Where an expectation is taken, it is with respect to the randomness in each of the pairs $(X^{(i)}, Y)$.

\subsection{Continuous case}
In the continuous setting we attempt to generate {\em joint} samples of the  pair $(X^{(i)}, Y)$.

Let $F_i : \mathcal{X}^{(i)} \times \mathcal{Y} \to \mathcal{Z}$ and $G_i : \mathcal{Z} \to \mathcal{X}^{(i)} \times \mathcal{Y}$ for $i = 1, ..., M$. We will refer to the maps $\{F_i : i = 1, ..., M\}$ as {\em encoders} and the maps $\{G_i : i = 1, ..., M\}$ as {\em decoders}. Then for each $j$, $W_j = F_j(X^{(j)}, Y)$ is a random variable taking values in the latent space $\mathcal{Z}$ and we define the random variable $Z_i$ to be a mixture of the random variables $\{W_j : j \neq i\}$ with $\mathbb{P}(Z_i = W_j) = \alpha_{ij}$. Sampling from $Z_i$ therefore corresponds to sampling uniformly from $\bigcup_{j \neq i} \mathcal{D}_j$ and then applying the corresponding $F_j$.

For $i = 1, ..., M$, we define the random variable $(\hat{X}^{(i)}, \hat{Y}) = G_i(Z_i) \in \mathcal{X}^{(i)} \times \mathcal{Y}$. We jointly train the maps $G_i, F_i$ for all $i$ simultaneously by introducing $M$ discriminators $D_1, ..., D_M$, with $D_i : \mathcal{X}^{(i)} \times \mathcal{Y} \to \mathbb{R}$, that (as in the standard GAN framework) attempt to distinguish real samples, $(X^{(i)}, Y)$, from fake samples, $(\hat{X}^{(i)}, \hat{Y})$.

We define the $i$th adversarial loss in this case to be
\begin{align}
\mathcal{L}_{adv}^i &= \mathbb{E}[\log D_i(X^{(i)}, Y)] + \mathbb{E}[\log (1-D_i(\hat{X}^{(i)}, \hat{Y}))]\nonumber\\
&= \mathbb{E}[\log D_i(X^{(i)}, Y)] + \mathbb{E}[\log (1-D_i(G_i(Z_i))]\nonumber\\
&= \mathbb{E}[\log D_i(X^{(i)}, Y)]\nonumber \\
&+ \sum_{j \neq i} \alpha_{ij} \mathbb{E}[\log(1-D_i(G_i(F_j(X^{(j)}, Y))))].
\end{align}
Note that $\mathcal{L}_{adv}^i$ depends on $D_i, G_i$ and $\{F_j : j \neq i\}$.

\begin{figure}[t!]
	\centering
	\includegraphics[width=0.5\textwidth]{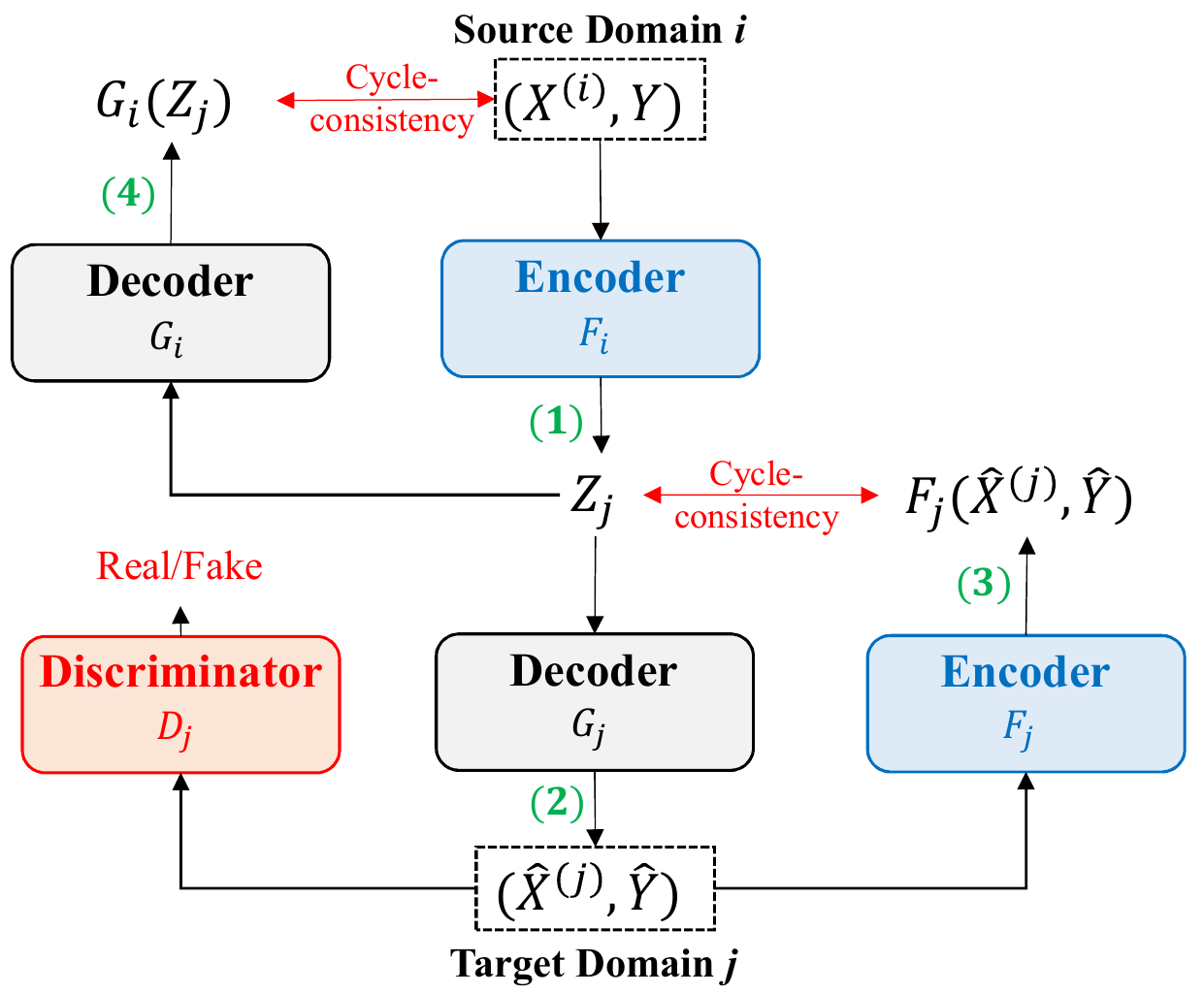}
	\caption{Block Diagram of RadialGAN - (1) $(X^{(i)},Y)$ in source domain $i$ is mapped to latent space by encoder $F_i$, (2) $Z_j$ in latent space is mapped to target domain $j$ by decoder $G_j$, (3) translated tuple $(\hat{X}^{(j)}, \hat{Y})$ is mapped back to latent space $Z$ using encoder $F_j$ to ensure cycle-consistency with $Z_j$, (4) $Z_j$ is mapped back to domain $i$ using decoder $G_i$ to ensure cycle-consistency with $(X^{(i)},Y)$.}
	\label{fig:block_diagram}
\end{figure}

We also introduce a cycle-consistency loss that ensures that translating into the latent space and back again returns something close to the original input and that mapping from the latent space into one of the domains and back again also returns something close to the original input, i.e. $G_i(F_i(x, y)) \approx (x, y)$ and $F_i(G_i(z)) \approx z$. This ensures that the encoding of each space into the latent space captures all the information present in the original space. Note that this also implies that translating into one of the other domains and translating back will also return the original input. We define the $i$th cycle-consistency loss as
\begin{align} \label{eq:z2}
\mathcal{L}_{cyc}^i &= \mathbb{E}[||(X^{(i)}, Y) - G_i(F_i(X^{(i)}, Y))||_2]\\\nonumber
&+ \mathbb{E}[||Z_i - F_i(G_i(Z_i))||_2]\\
&= \mathbb{E}[||(X^{(i)}, Y) - G_i(F_i(X^{(i)}, Y))||_2]\\\nonumber
&+ \sum_{j \neq i} \alpha_{ij}\mathbb{E}[||F_j(X^{(j)}, Y) - F_i(G_i(F_j(X^{(j)}, Y)))||_2]
\end{align}
where $||\cdot||_2$ is the standard $\ell_2$-norm. Note that $\mathcal{L}_{cyc}^i$ depends on $G_i$ and $\{F_j : j = 1, ..., M\}$.

\subsection{Discrete case}
In the discrete setting, rather than attempting to generate the label, $Y$, we instead create additional samples by {\em conditioning} on $Y$ and generating according to the distributions $X^{(i)} | Y = y$ for each $i = 1, ..., M, y \in \mathcal{Y}$. Motivated by the conditional GAN framework \cite{conditionalgan}, \cite{acgan}, we do this by passing $Y$ as an input to our generator.

Let $F_i : \mathcal{X}^{(i)} \times \mathcal{Y} \to \mathcal{Z}$ and $G_i : \mathcal{Z} \times \mathcal{Y} \to \mathcal{X}^{(i)}$ for $i = 1, ..., M$. Note that now $G_i$ takes $Y$ as an input, i.e. it conditions on $Y$. As before, for each $j$, we define the random variable $W_j$ by $W_j = F_j(X^{(j)}, Y) \in \mathcal{Z}$ and we similarly define $Z_i$ as above.

In contrast to the continuous case, we now define $\hat{X}^{(i)} = G_i(Z_i, Y)$ and the discriminators, $D_i$, are now trying to distinguish real samples of $X^{(i)} | Y = y$ from fake samples $\hat{X}^{(i)} | Y = y$.

The adversarial and cycle-consistency losses are defined in a similar manner to the continuous case.

\subsection{Training}
For the remainder of this section, we will focus on the continuous case, but equivalent expressions can be written for and similar discussions apply to the discrete case. For all $i$, we implement each of $F_i$, $G_i$ and $D_i$ as multi-layer perceptrons and denote their parameters by $\theta_F^i$, $\theta_G^i$ and $\theta_D^i$ respectively.

Using the losses $\mathcal{L}_{adv}^i$ and $\mathcal{L}_{rec}^i$ we define the objective of RadialGAN as the following minimax problem
\begin{align}
\min_{\mathbf{G}, \mathbf{F}} \max_{\mathbf{D}} \biggl(\sum_{i = 1}^M \mathcal{L}_{adv}^i(D_i, G_i&, \{F_j : j \neq i\})\\\nonumber
&+ \lambda \sum_{i = 1}^M \mathcal{L}_{cyc}^i(G_i, \mathbf{F})\biggr)
\end{align}
where $\mathbf{G} = (G_1, ..., G_M), \mathbf{F} = (F_1, ..., F_M)$ and $\mathbf{D} = (D_1, ..., D_M)$ and $\lambda > 0$ is a hyper-parameter.

As in the standard GAN framework, we solve this minimax problem iteratively by first training $\mathbf{D}$ with a fixed $\mathbf{G}$ and $\mathbf{F}$ using a mini-batch of size $k_D$ and then training $\mathbf{G}$ and $\mathbf{F}$ with a fixed $\mathbf{D}$ using a mini-batch of size $k_G$. Pseudo-code for our algorithm can be found in Algorithm \ref{alg:pseudo}. Fig. \ref{fig:block_diagram} depicts the interactions that occur between the $F_i, G_i, F_j, G_j$ and $D_j$ for $i \neq j$ when $Z_j = W_i$.

\subsection{Prediction}
After we have trained the translation functions $G_1, ..., G_M$ and $F_1, ..., F_M$, we create $M$ {\em augmented} datasets $\mathcal{D}_1', ..., \mathcal{D}_M'$ where $\mathcal{D}_i' = \mathcal{D}_i \cup \bigcup_{j \neq i} G_i(F_j(\mathcal{D}_j))$. The predictors $f_1, ..., f_M$ are then learned on these augmented datasets. In Section \ref{sect:experiments} we model each $f_i$ as a logistic regression, but we stress that our method can be used for any choice of predictive model and we show further results in the Supplementary Materials where we model each $f_i$ as a multi-layer perceptron to demonstrate this point.

\begin{algorithm}[t!]
	\caption{Pseudo-code of RadialGAN}
	\label{alg:pseudo}
	\begin{algorithmic}
		\STATE {\bf Initialize: }$\theta_G^1, ..., \theta_G^M, \theta_F^1, ..., \theta_F^M, \theta_D^1, ..., \theta_D^M$ 
		\WHILE{training loss has not converged}
		\STATE \textbf{(1) Update $\mathbf{D}$ with fixed $\mathbf{G}$, $\mathbf{F}$}
		\FOR{$i = 1, ..., M$}
		\STATE Draw $k_D$ samples from $\mathcal{D}_i$, $\{(x^{(i)}_k, y^i_k)\}_{k = 1}^{k_D}$
		\STATE Draw $k_D$ samples from $\bigcup_{j \neq i} \mathcal{D}_j$, $\{(x^{(j_k)}_k, y_k)\}_{k=1}^{k_D}$
		\FOR{$k = 1, ..., k_D$}
		\STATE
		\begin{equation*}
		(\hat{x}^{(i)}_k, \hat{y}_k) \gets G_i(F_{j_k}(x^{(j_k)}_k, y_k))
		\end{equation*}
		\ENDFOR
		\STATE Update $\theta_D^i$ using stochastic gradient descent(SGD)
		\begin{align*}
		\triangledown_{\theta_D^i} - \biggl(\sum_{k = 1}^{k_D}& \log D_i(x^{(i)}_k, y^i_k)\\
		&+ \sum_{k = 1}^{k_D} \log(1 - D_i(\hat{x}^{(i)}_k, \hat{y}_k))\biggr)
		\end{align*}
		\ENDFOR
		\STATE \textbf{(2) Update $\mathbf{G}$, $\mathbf{F}$ with fixed $\mathbf{D}$}
		\FOR{$i = 1, ..., M$}
		\STATE Draw $k_G$ samples from $\mathcal{D}_i$, $\{(x^{(i)}_k, y^i_k)\}_{k = 1}^{k_G}$
		\STATE Draw $k_G$ samples from $\bigcup_{j \neq i} \mathcal{D}_j$, $\{(x^{(j_k)}_k, y_k)\}_{k=1}^{k_G}$
		\ENDFOR
		\STATE Update $\theta_{\mathbf{G, F}} = (\theta_G^1, ..., \theta_G^M, \theta_F^1, ..., \theta_F^M)$ using SGD
		\begin{align*}
		&\triangledown_{\theta_{\mathbf{G, F}}} \sum_{i = 1}^M \sum_{k = 1}^{k_G} \log (1 - D_i(G_i(F_{j_k}(x^{(j_k)}_k, y_k)))) +\\
		& \lambda \sum_{i = 1}^M \sum_{k = 1}^{k_G} ||(x^{(i)}_k, y_k) - G_i(F_i(x^{(i)}_k, y_k))||_2+ \\
		& \lambda \sum_{i = 1}^M \sum_{k = 1}^{k_G} ||F_{j_k}(x^{(j_k)}_k, y_k) - F_i(G_i(F_{j_k}(x^{(j_k)}_k, y_k)))||_2
		\end{align*}
		\ENDWHILE
	\end{algorithmic}
\end{algorithm}

\begin{figure*}[t!]
	\centering
	\begin{minipage}{0.18\textwidth}
		\centering
		\includegraphics[width=\textwidth]{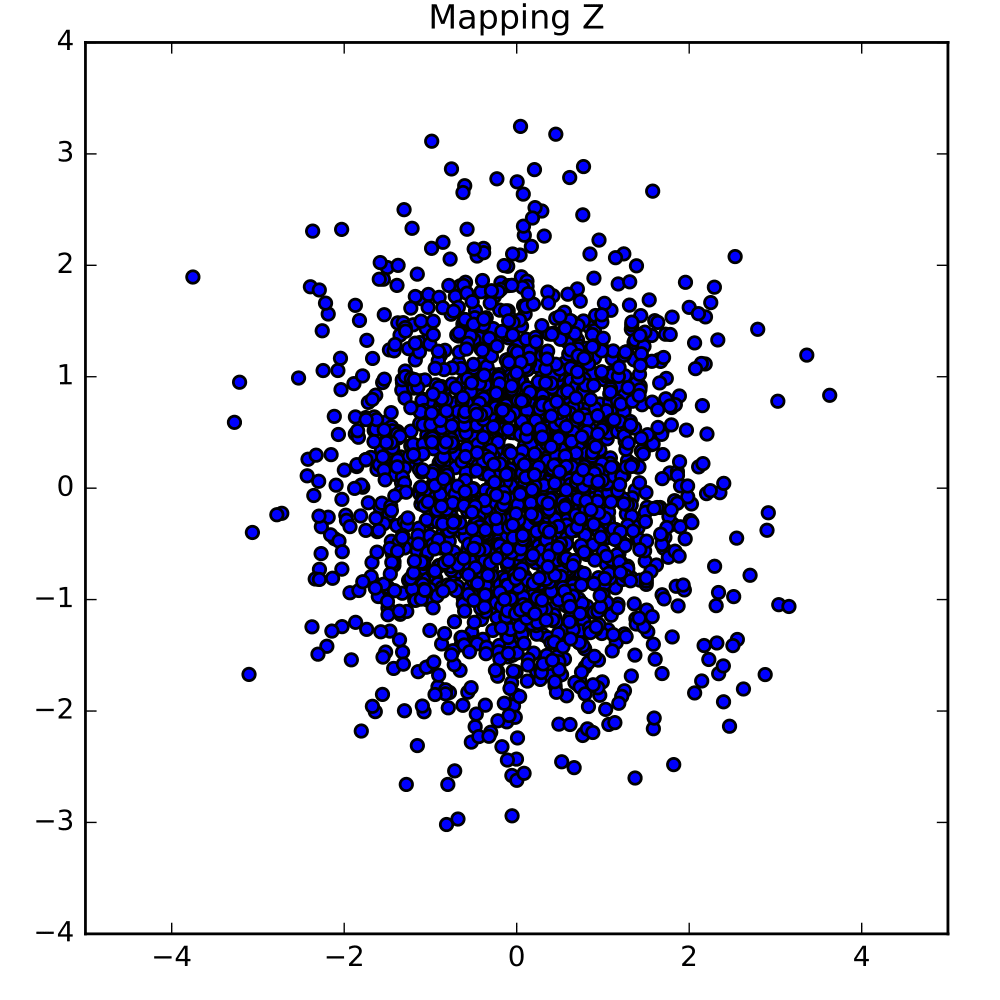}%
	\end{minipage}
	\textcolor{black}{\vrule width 2pt}
	\begin{minipage}{0.18\textwidth}
		\centering
		\includegraphics[width=\textwidth]{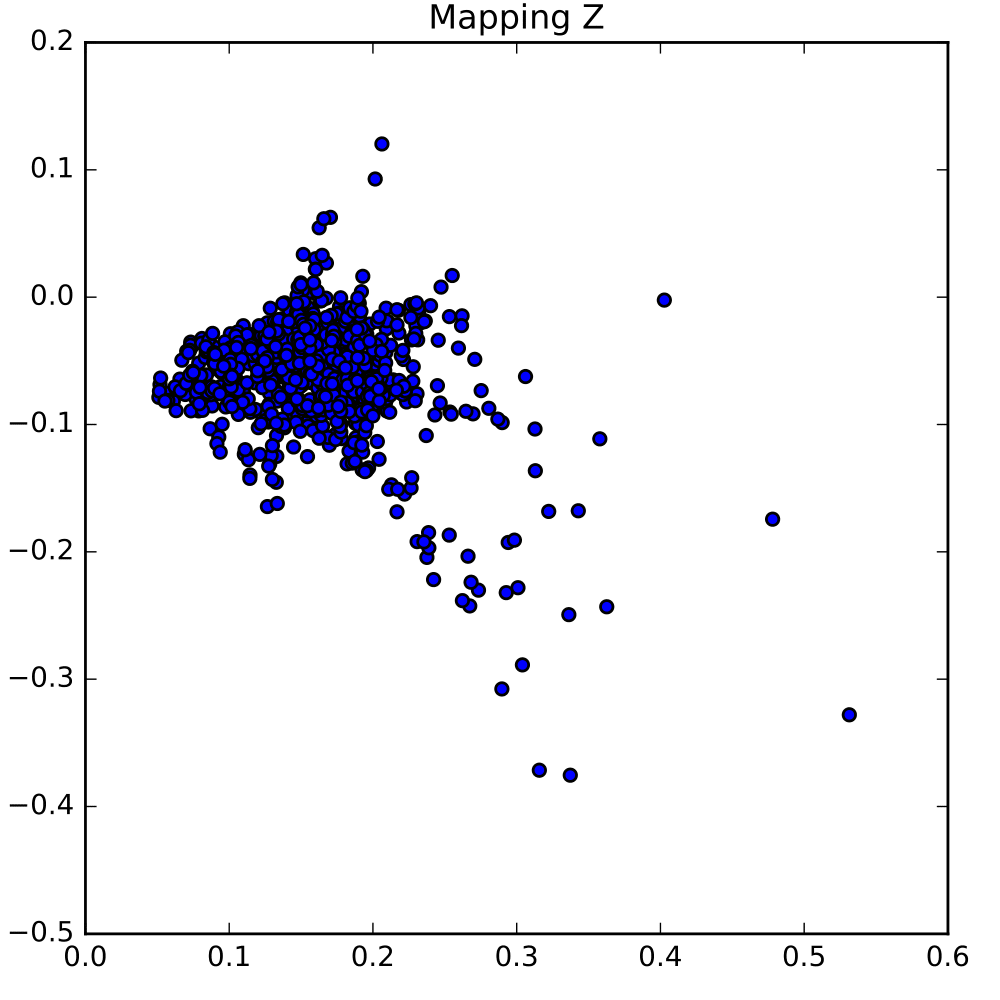}
	\end{minipage}
	\begin{minipage}{0.18\textwidth}
		\centering
		\includegraphics[width=\textwidth]{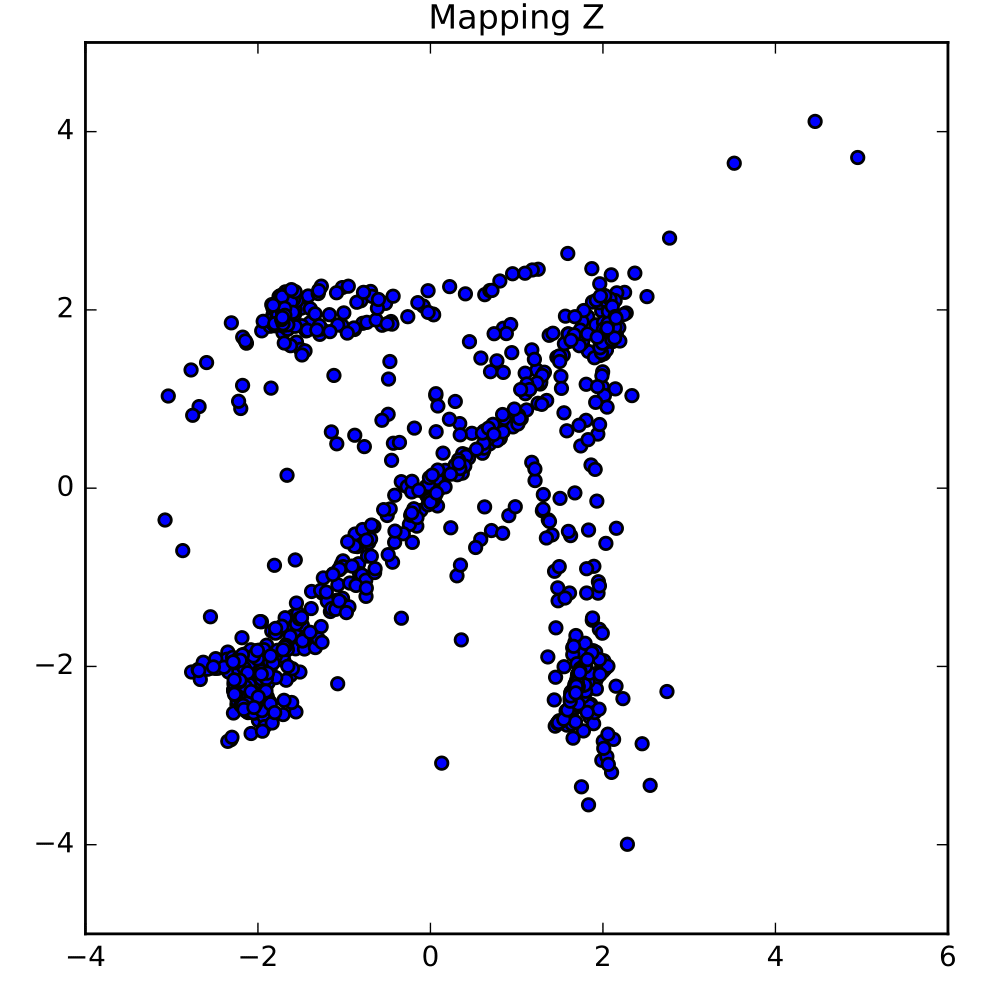}
	\end{minipage}
	\begin{minipage}{0.18\textwidth}
		\centering
		\includegraphics[width=\textwidth]{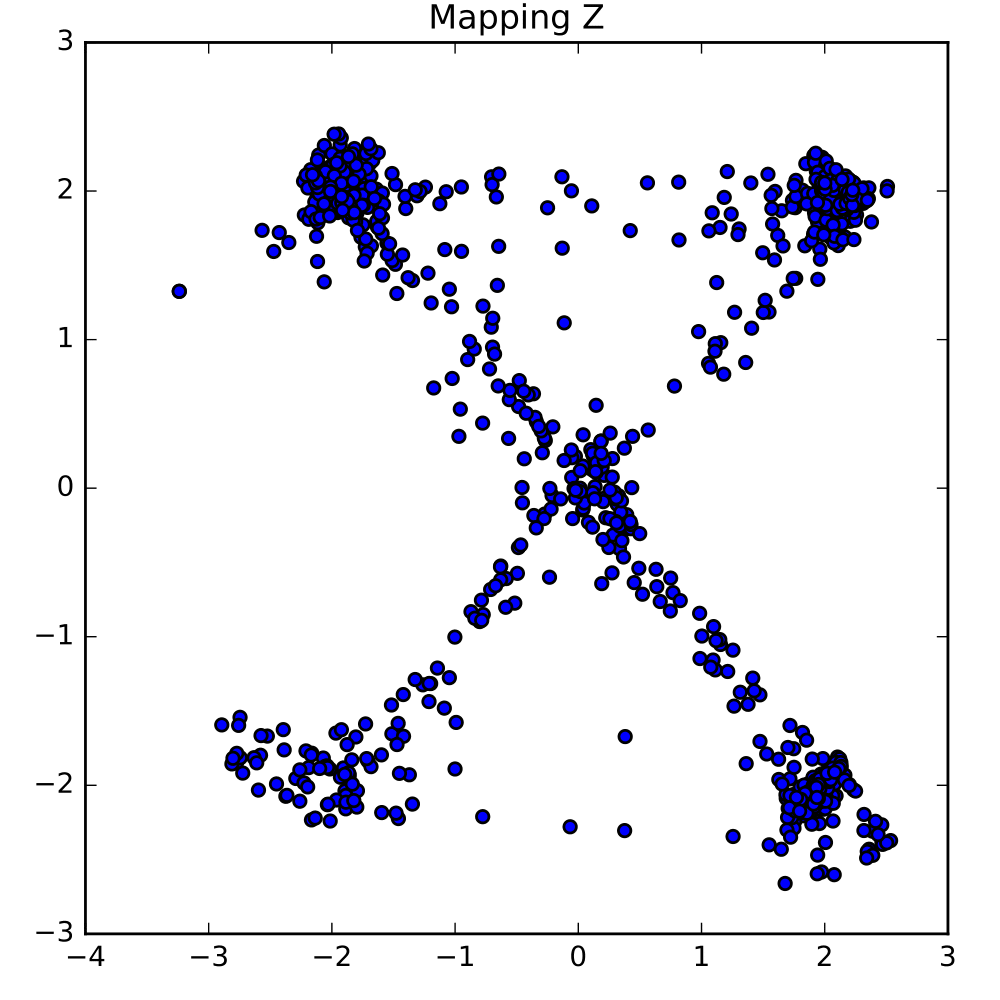}
	\end{minipage}
	\textcolor{black}{\vrule width 2pt}
	\begin{minipage}{0.18\textwidth}
		\centering
		\includegraphics[width=\textwidth]{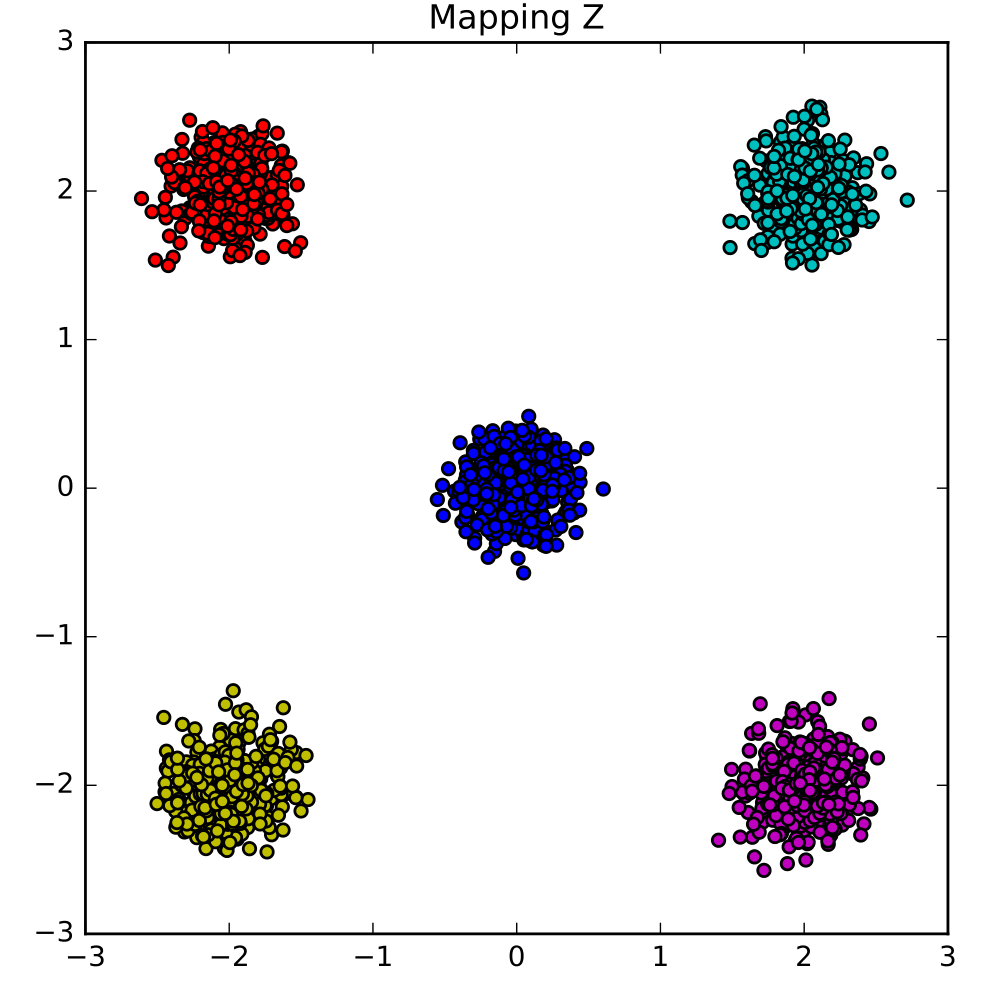}%
	\end{minipage}
\end{figure*}

\begin{figure*}[t!]
	\centering
	\begin{minipage}{0.18\textwidth}
		\centering\includegraphics[width=\textwidth]{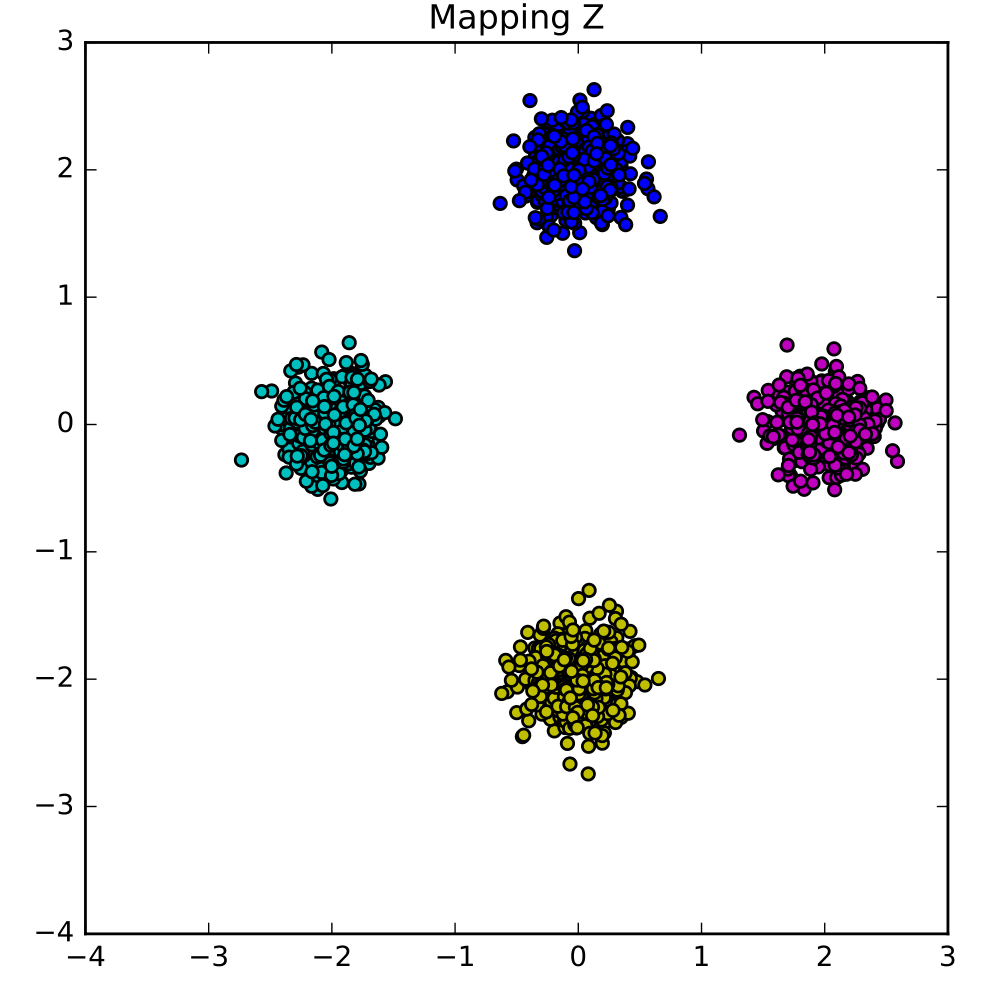}%
		\captionsetup{labelformat=empty}
		\caption*{(a) Initial distribution}
	\end{minipage}
	\textcolor{black}{\vrule width 2pt}
	\begin{minipage}{0.18\textwidth}
		\centering
		\includegraphics[width=\textwidth]{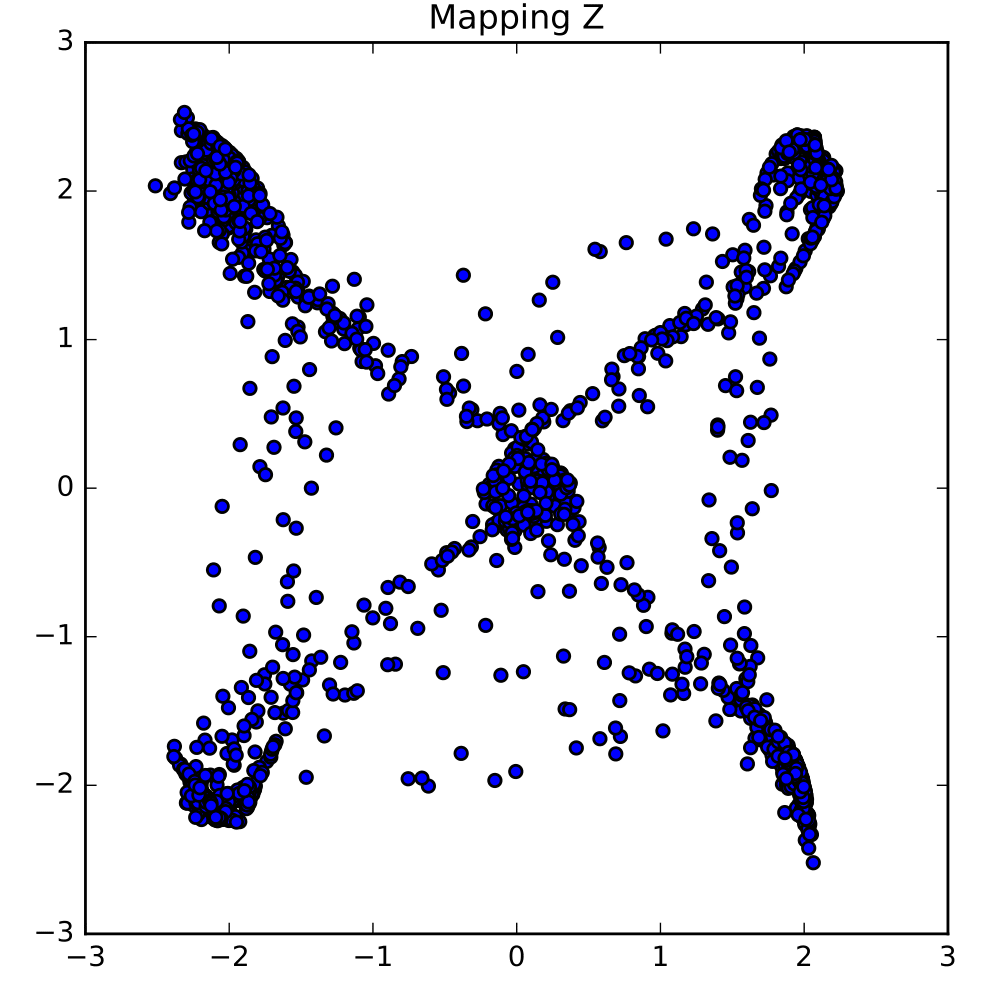}
		\captionsetup{labelformat=empty}
		\caption*{(b) $n=300$}
	\end{minipage}
	\begin{minipage}{0.18\textwidth}
		\centering
		\includegraphics[width=\textwidth]{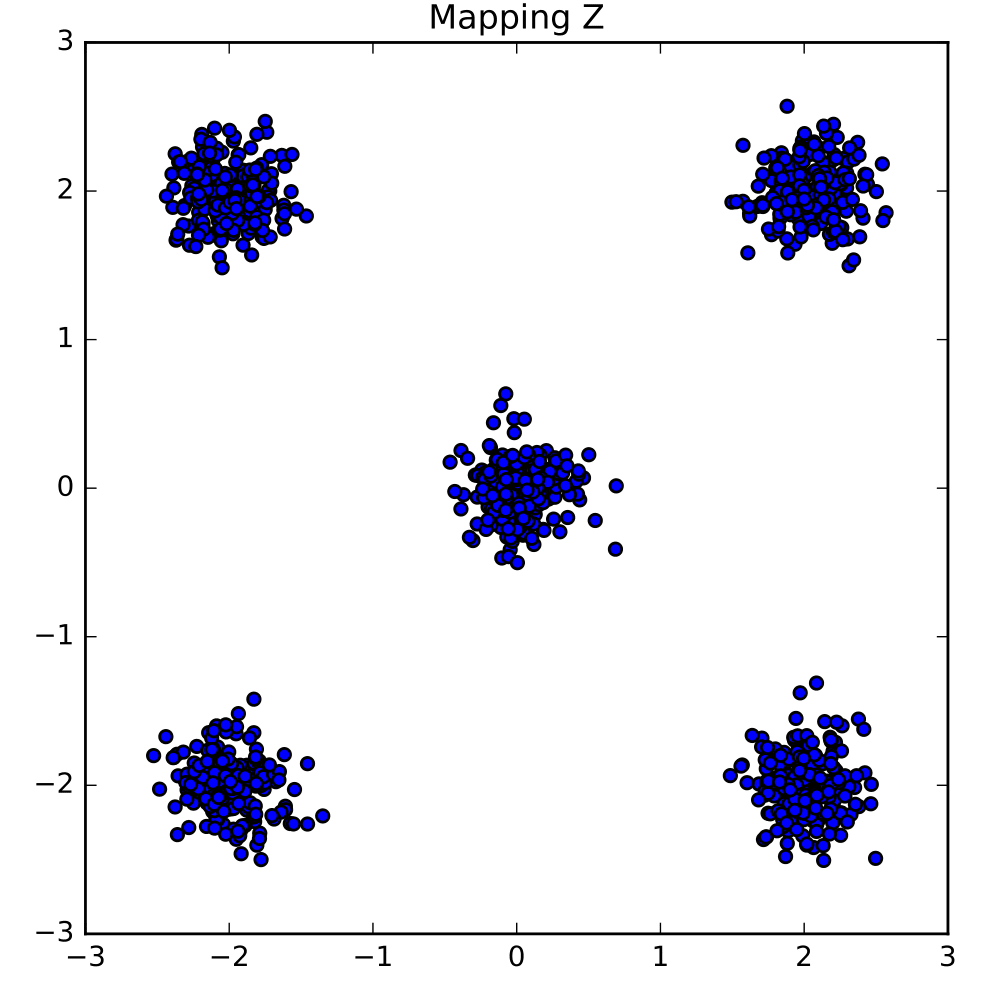}
		\captionsetup{labelformat=empty}
		\caption*{(c) $n=500$}
	\end{minipage}
	\begin{minipage}{0.18\textwidth}
		\centering
		\includegraphics[width=\textwidth]{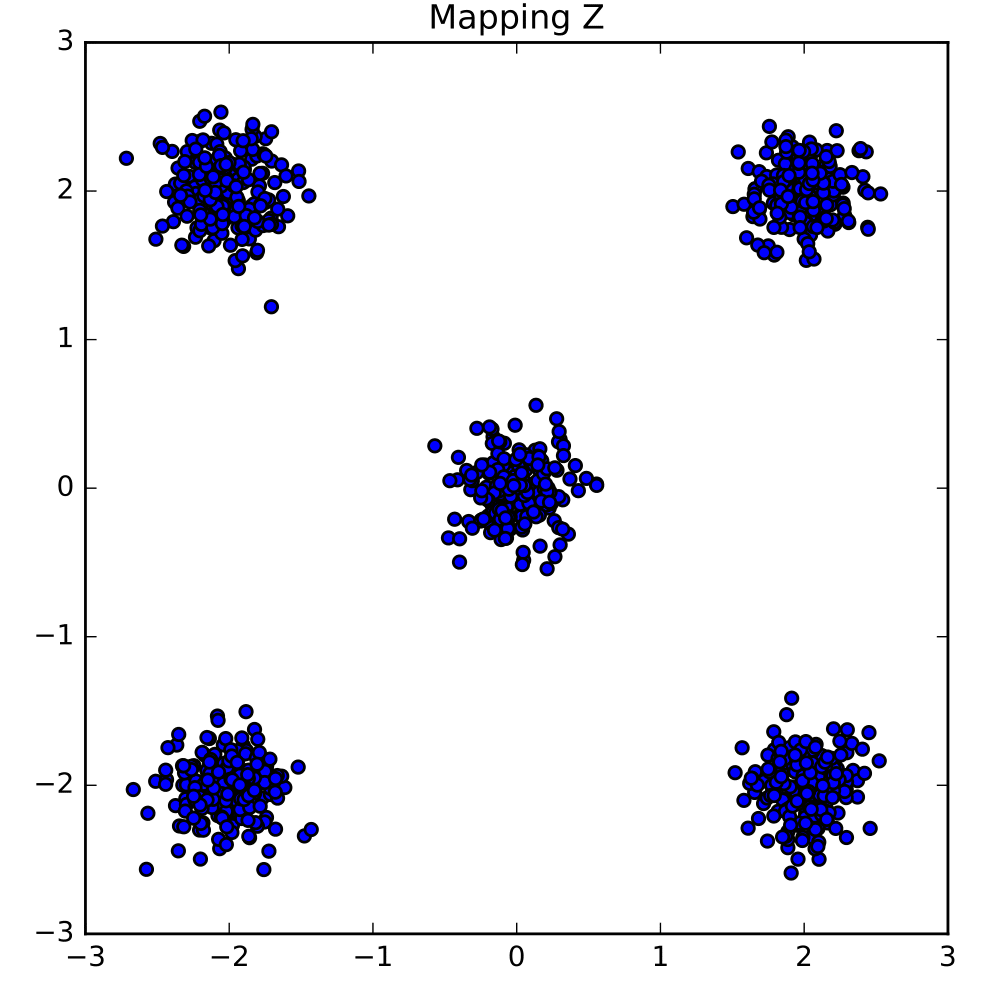}
		\captionsetup{labelformat=empty}
		\caption*{(d) $n=1000$}
	\end{minipage}
	\textcolor{black}{\vrule width 2pt}
	\begin{minipage}{0.18\textwidth}
		\centering
		\includegraphics[width=\textwidth]{target.png}%
		\captionsetup{labelformat=empty}
		\caption*{(e) Target distribution}
	\end{minipage}
	\caption{Mapping functions between two distributions with different initial distributions (upper: normal, lower: Gaussian mixtures).}
	\label{fig:similar_dist}
\end{figure*}

\subsection{Improved learning using WGAN}
The model proposed above is cast in the original GAN setting outlined by \cite{standardgan}. To improve the robustness of the GAN training, we can instead cast the model in the WGAN setting \cite{improvedwgan} by replacing $\mathcal{L}_{adv}^i$ with the loss proposed given by
\begin{align}
\mathcal{L}_{wgan}^i &= \mathbb{E}[D_i(X^{(i)})] - \mathbb{E}[D_i(G_i(Z_i))] \\\nonumber
&\qquad+ \beta \mathbb{E}[(||\triangledown_{x}D_i(\bar{X}_i)||_2 -1)^2]\\\nonumber
&= \mathbb{E}[D_i(X^{(i)})] - \sum_{j \neq i} \alpha_{ij} \mathbb{E}[D_i(G_i(F_j(X^{(j)})))]\\\nonumber
&\qquad+ \beta \mathbb{E}[(||\triangledown_{x}D_i(\bar{X}_i)||_2 -1)^2]
\end{align}
where $\bar{X}$ is given by sampling uniformly along straight lines between pairs of samples of $X^{(i)}$ and $\hat{X}^{(i)}$ and $\beta$ is a hyper-parameter.

\section{Experiments}\label{sect:experiments}
\subsection{Verifying intuition}
Using a synthetic example, we provide further experimental results to back up the intuition outlined in Section \ref{sec:motiv} and in Fig. \ref{fig:motivation1}. In this experiment we show that if the initial and target distributions are similarly shaped, the GAN framework requires fewer samples in order to learn a mapping between them.

Fig. \ref{fig:similar_dist} depicts the results of this experiment in which the target distribution is a Gaussian mixture with 5 modes, showing a comparison between the quality of the mapping function when the initial distribution is a simple Gaussian distribution (top row of figure) compared to a Gaussian mixture with 4 modes (bottom row of figure). As can be seen in Fig. \ref{fig:similar_dist}, learning a good mapping from the 4 mode Gaussian mixture to the target distribution requires fewer samples than learning a good mapping from the simple Gaussian distribution to the target.

\begin{table*}[t!]
	\renewcommand{\arraystretch}{1.2}
	\caption{Prediction performance comparison with different number of datasets (Red: Negative effects)}
	\label{tab:Prediction}
	\centering
    \small
	\begin{tabular}{|c|c|c|c|c|c|c|}
		\hline
		\multirow{2}{*}{\textbf{Algorithm}}&\multicolumn{2}{c|}{\textbf{$M=3$}} &\multicolumn{2}{c|}{\textbf{$M=5$}} &\multicolumn{2}{c|}{\textbf{$M=7$}} \\
		\cline{2-7}
		& \textbf{AUC} & \textbf{APR} & \textbf{AUC} & \textbf{APR}& \textbf{AUC} & \textbf{APR} \\
		\hline
		\textbf{RadialGAN} & .0154$\pm$.0091 & .0243$\pm$.0096  & .0292$\pm$.0009 & .0310$\pm$.0096 & .0297$\pm$.0071 & .0287$\pm$.0073 \\
		\hline
		{Simple-combine} & .0124$\pm$.0020 & .0110$\pm$.0016  & {{.0132$\pm$.0020}} & {{.0118$\pm$.0026}} & .0135$\pm$.0017 & .0156$\pm$.0025 \\
		\hline
		{Co-GAN} & .0058$\pm$.0028 & {.0085$\pm$.0026} & .0094$\pm$.0018 & .0139$\pm$.0036 & {\color{red}{-.0009$\pm$.0015}} & {\color{red}{-.0013$\pm$.0027}}  \\
		\hline
		{StarGAN} & .0119$\pm$.0015 & .0150$\pm$.0013 & .0150$\pm$.0025 & .0191$\pm$.0013  &  .0121$\pm$.0020 & .0160$\pm$.0021 \\
		\hline
		{Cycle-GAN} & {\color{red}{-.0228$\pm$.0112}}  & {\color{red}{-.0306$\pm$.0085}}  & {\color{red}{-.0177$\pm$.0082}} & {\color{red}{-.0196$\pm$.0085}} & {\color{red}{-.0076$\pm$.0022}} & {\color{red}{-.0168$\pm$.0030 }} \\
		\hline
		{\cite{sameproblem}} & {\color{red}{-.0314$\pm$.0075}} & {\color{red}{-.0445$\pm$.0125}}  & {\color{red}{-.0276$\pm$.0057}} & {\color{red}{-.0421$\pm$.0052}} & {\color{red}{-.0292$\pm$.0054}} & {\color{red}{-.0411$\pm$.0063}}  \\
		\hline
	\end{tabular}
\end{table*}

\subsection{Experiment Setup}
\textbf{Data Description: }The remainder of the experiments in this section are all performed using real-world datasets. MAGGIC \cite{maggic} is a collection of 30 different datasets from 30 different medical studies containing patients that experienced heart failure. Among the 30 datasets, we use the 14 studies that each contain more than 500 patients to allow us to create a large test set for each dataset. We set the label of each patient as 1-year all-cause mortality, excluding all patients who are censored before 1 year. Across the 14 selected studies the total number of features is 216, with the average number of features in a single study being 66. There are 35 features (53.8\%) shared between any two of the selected studies on average. The average number of patients in each selected study is 3008 with all of the selected studies containing between 528 and 13279 patients.\footnote{More details of the datasets (including study-specific statistics) can be found in the Supplementary Materials.} For each dataset, we randomly sample 300 patients to be used as training samples and the remaining samples are used for testing.

\textbf{Benchmarks: }We demonstrate the performance of RadialGAN against 6 benchmarks. In the Target-only benchmark, we use just the target dataset to construct a predictive model. In the Simple-combine benchmark, we combine all the datasets by defining the feature space to be the union of all feature spaces, treating unmeasured values as missing and setting them to zero. We also concatenate the mask vector to capture the missingness of each feature and then learn a predictive model on top of this dataset. The conditional-GAN (Co-GAN) and StarGAN (StarGAN) benchmarks also use this combined dataset. The Cycle-GAN benchmark learns pairwise translation functions between pairs of datasets (rather than mapping through a central latent space). For a given target dataset \cite{sameproblem} creates an augmented dataset by taking the additional dataset, discarding features not present in the target dataset, and augmenting with $0$s to account for features present in the target but not the source. The predictive model is then learned on this augmented dataset. The way of hyper-parameter optimization is explained in the Supplementary Materials.

\textbf{Metrics: }As the end goal is prediction, not domain translation, we use the prediction performance of a logistic regression and a 2-layer perceptron to measure the quality of the domain translation algorithms. We report two types of prediction accuracy: Area Under ROC Curve (AUC) and Average Precision Recall (APR)\footnote{Unless otherwise stated, we calculate these individually for each dataset and then report the average across all datasets.}\footnote{The results presented in this section are for logistic regression, with the 2-layer perceptron results reported in the Supplementary Materials.}. Furthermore, we set the performance of the Target-only predictive model as the baseline and report the performance gain of other benchmarks (including RadialGAN) over the Target-only predictive model. We run each experiment 10 times with 5-fold cross validation each time and report the average performance over the 10 experiments.

\subsection{Comparison with Target-only GAN}
Before extensively comparing RadialGAN with the previously mentioned benchmarks, we first demonstrate the advantage of using auxiliary datasets to generate additional samples over using just the target dataset to train a generative model. We compare the performance of a predictive model trained on 3 different datasets: the original target dataset (Target-only), one enlarged by translating other datasets to the target domain using RadialGAN, and one enlarged using a standard GAN trained to simulate the target dataset (and using no other datasets, i.e. only use the target dataset).

\begin{figure}[t!]
	\centering
	\includegraphics[width=0.48\textwidth]{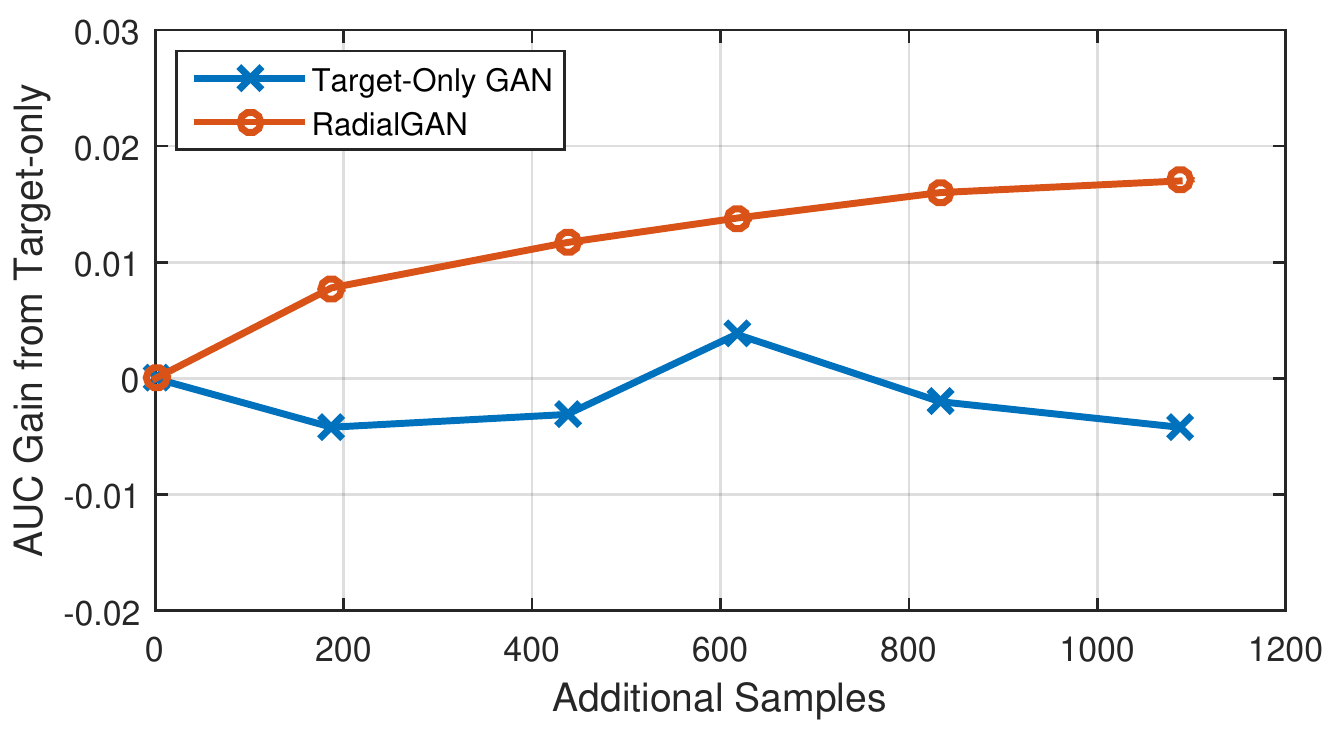}
	\caption{Comparison to Target-Only GAN}
	\label{fig:target_only_gan}
\end{figure}

As can be seen in Fig. \ref{fig:target_only_gan}, as the number of additional samples we provide increases, RadialGAN achieves higher prediction gains over the target-only predictive model. On the other hand, the target-only GAN shows a tendency to worsen the predictive model, particularly when a larger number of samples are generated. This is because the samples generated by a target-only GAN cannot possibly include information that is not already contained in the dataset used to train the GAN (i.e. the target dataset). Moreover, due to the limited size of the target dataset, the samples produced by the GAN will be noisy, and so when a large number of them are generated, the resulting dataset can be significantly noisier than the original. On the other hand, RadialGAN is able to leverage the information contained across multiple datasets to generate higher quality (and therefore less noisy) additional samples.

\begin{figure*}[t!]
	\centering
	\includegraphics[width=0.9\textwidth]{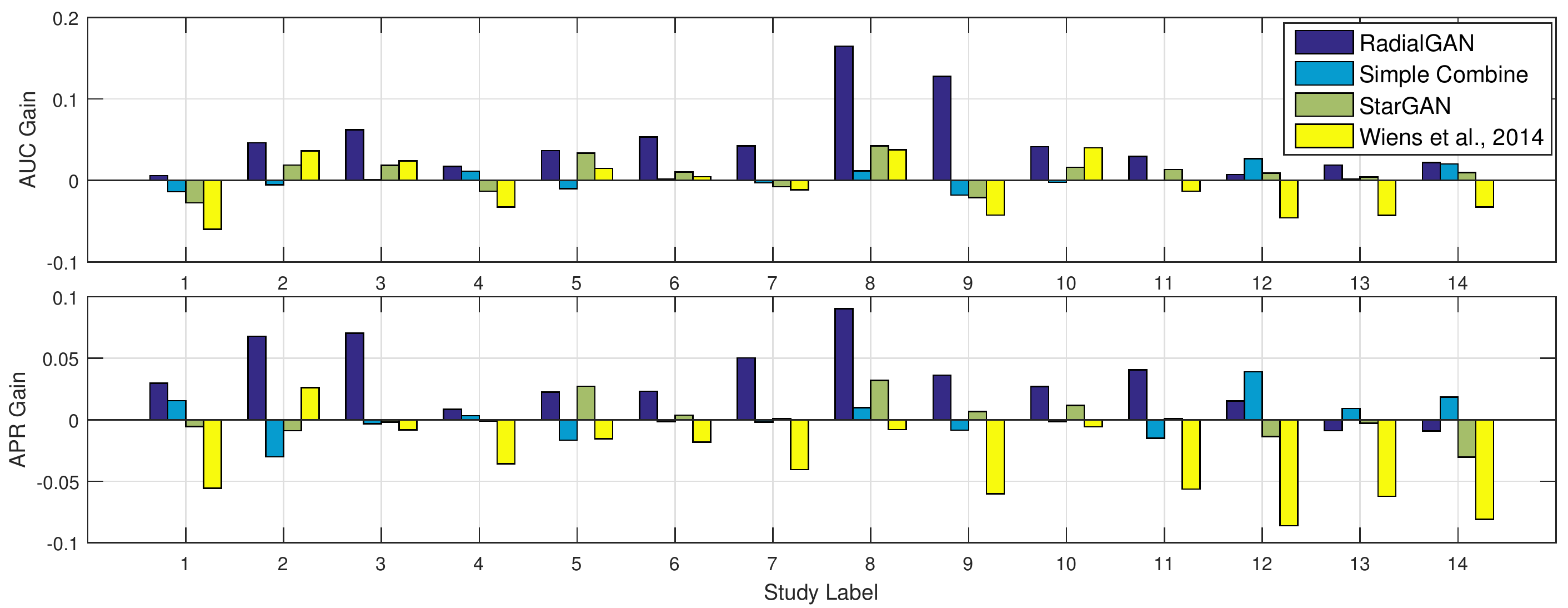}
	\caption{Performance improvements on each of the 14 studies.}
	\label{fig:all_perform}
\end{figure*}

\subsection{Utilizing Multiple Datasets}
In this subsection, we compare RadialGAN to the other benchmarks when we vary the total number of datasets. In the first experiment we select $3, 5$ or $7$ datasets randomly from among the $14$ already chosen. Table \ref{tab:Prediction} shows the average improvements of prediction performance over the Target-only benchmark.

As can be seen in Table \ref{tab:Prediction}, RadialGAN consistently leads to more accurate target-specific predictive models than the target-only predictive model and other benchmarks. Moreover, as the number of domains (datasets) increases (from $M=3$ to $M=7$), the performance gain increases due to the larger number of samples that RadialGAN can utilize. The performance of \cite{sameproblem} is usually worse than the Target-only model due to the fact that it does not address the problem of distribution mismatch across the domains. Simple-combine, Co-GAN, and StarGAN each treat the problem in a way that requires learning from a large sparse matrix (equivalent to their being a large amount of missing data); thus, even though the number of available samples increases, the improvement of prediction performance is marginal due to the high dimensional data.

\textbf{Utilizing all datasets: }In this experiment, we set $M=14$ and so use all 14 datasets. We report the improvement of AUC and APR for each study over the Target-only baseline for each study individually in Fig. \ref{fig:all_perform}. We compare RadialGAN with three competitive benchmarks (Simple-combine, StarGAN and \cite{sameproblem}). As can be seen in Fig. \ref{fig:all_perform}, RadialGAN outperforms the Target-only predictive model and other benchmarks in almost every study. Furthermore, significant improvements can be seen in some cases (such as Study 8 and 9). On the other hand, the performance improvements by other benchmarks are marginal in all cases and often the performance is decreased due to the introduction of noisy additional samples.

\subsection{Two extreme cases}
In this paper, we address two challenges of transfer learning: (1) feature mismatch, (2) distribution mismatch. To understand the source of gain, we conduct two further experiments. First, we evaluate the performance of RadialGAN in a setting where there is no feature mismatch (i.e. the datasets all contain the same features), which we call Setting A. We use the same MAGGIC dataset (used in Table \ref{tab:Prediction}) with $M=5$ but only use the features that are shared across all 5 datasets. Second, we evaluate the performance of RadialGAN in a setting where there is no distribution mismatch, which we call Setting B. For this, we use the study in the original MAGGIC dataset with the most samples (13,279 patients) and randomly divide it into 5 subsets. Then, we randomly remove 33\% of the features in each subset; thus, introducing a different feature set for each subset. 

\begin{table}[t!]
	\renewcommand{\arraystretch}{1.2}
	\caption{Prediction performance comparison without feature mismatch (Setting A) or distribution mismatch (Setting B)}
	\label{tab:extreme_case}
	\centering
	\begin{tabular}{|c|c|c|c|c|}
		\hline
		\multirow{2}{*}{\textbf{Algorithm}}&\multicolumn{2}{c|}{Setting A} &\multicolumn{2}{c|}{Setting B}  \\
		\cline{2-5}
		& \textbf{AUC} & \textbf{APR} & \textbf{AUC} & \textbf{APR} \\
		\hline
		\textbf{RadialGAN}& .0169 & .0249& .0313 & .0331 \\
		\hline
		{Simple-combine}& .0098&.0183 & .0271 & .0285\\
		\hline
		{Co-GAN}&.0027 &.0173 & {\color{red}{-.0029}} & {\color{red}{-.0013}}\\
		\hline
		{StarGAN}& .0084&.0142 & .0020 & .0042\\
		\hline
		{Cycle-GAN}& {\color{red}{-.0113}}& {\color{red}{-.0107}}& .0045 & .0031\\
		\hline
		{\cite{sameproblem}}&{\color{red}{-.0148}} &{\color{red}{-.0172}} & .0249 & .0299\\
		\hline
	\end{tabular}
\end{table}

As can be seen in Table \ref{tab:extreme_case}, the performance of RadialGAN is competitive with the other benchmarks in both Settings. The performance gain is smaller than can be seen in Table \ref{tab:Prediction} due to the fact that RadialGAN is the only algorithm designed to efficiently deal with both challenges, though RadialGAN is still competitive in both settings individually. Note that \cite{sameproblem} works well in Setting B because it is designed for the setting where there is no distribution mismatch. On the other hand, the performance gain of StarGAN decreases in Setting B because it does not naturally address the challenge of feature mismatch.

\section{Conclusion}
In this paper we proposed a novel approach for utilizing multiple datasets to improve the performance of target-specific predictive models. Future work will investigate methods for determining which datasets should or should not be included in a specific analysis. By doing so, we hope to build on the current algorithm such that it can automatically learn to include suitable datasets which result in an improvement.

\newpage
\section*{Acknowledgement}
The authors would like to thank the reviewers for their helpful comments. The research presented in this paper was supported by the Office of Naval Research (ONR) and the NSF (Grant number: ECCS1462245, ECCS1533983, and ECCS1407712).

\end{document}